\documentclass[conference]{IEEEtran}
\IEEEoverridecommandlockouts
\usepackage{cite}
\usepackage{amsmath,amssymb,amsfonts}
\usepackage{algorithmic}
\usepackage{graphicx}
\usepackage{textcomp}
\usepackage{xcolor}
\usepackage{subfig}
\def\BibTeX{{\rm B\kern-.05em{\sc i\kern-.025em b}\kern-.08em
    T\kern-.1667em\lower.7ex\hbox{E}\kern-.125emX}}
\begin{document}


\title{BottleNet++: An End-to-End Approach for Feature Compression in Device-Edge Co-Inference Systems
}

\author{\IEEEauthorblockN{Jiawei Shao, Jun Zhang}
\IEEEauthorblockA{Department of Electronic and Information Engineering \\
The Hong Kong Polytechnic University, Hong Kong\\
Email: jiawei.shao@connect.polyu.hk, jun-eie.zhang@polyu.edu.hk}
}

\maketitle

\begin{abstract}
The emergence of various intelligent mobile applications demands the deployment of powerful deep learning models at resource-constrained mobile devices. The device-edge co-inference framework provides a promising solution by splitting a neural network at a mobile device and an edge computing server. In order to balance the on-device computation and the communication overhead, the splitting point needs to be carefully picked, while the intermediate feature needs to be compressed before transmission. Existing studies decoupled the design of model splitting, feature compression, and communication, which may lead to excessive resource consumption of the mobile device. In this paper, we introduce an end-to-end architecture, named BottleNet++, that consists of an encoder, a non-trainable channel layer, and a decoder for more efficient feature compression and transmission. The encoder and decoder essentially implement joint source-channel coding via lightweight convolutional neural networks (CNNs), while explicitly considering the effect of channel noise. By exploiting the strong sparsity and the fault-tolerant property of the intermediate feature in deep neural networks (DNNs), BottleNet++ achieves a much higher compression ratio than existing methods. Compared with directly transmitting intermediate data without feature compression, BottleNet++ achieves up to 64$\times$ bandwidth reduction over the additive white Gaussian noise channel and up to 256$\times$ bit compression ratio in the binary erasure channel, with less than 2\% reduction in accuracy of classification.
\end{abstract}


\section{Introduction}

Recently, applications enabled by various mobile and Internet of Things (IoT) devices have profoundly changed our daily life\cite{iotlajihua1}. One primary driver of these applications is the recent breakthrough in Deep Neural Networks (DNNs) \cite{iot} that for reliable inference.
Unfortunately, DNN-based applications typically require a tremendous amount of computation, so they cannot be directly deployed on resource-constrained mobile/edge devices.
A common method to solve this problem is to transmit the raw data to be processed at the cloud or edge computing platforms\cite{cloud-based}, \cite{accel}.
The main disadvantage of this approach is the huge amount of communication overhead, which leads to considerable latency and energy consumption\cite{latencylajiahua1}. 
Another approach is on-device inference, which deploys compressed DNNs on mobile devices.
However, the over-compressed networks cause severe performance degradation, and the limited resources lead to high on-device communication latency.
These difficulties have driven the development of other alternatives, among which \textit{device-edge co-inference} is a promising solution\cite{Edge-host}, \cite{partitionlajiahua}.
This method splits a network into the front part on a mobile device and the remaining part on an edge server. The output of the front part network is the intermediate feature, which would be transmitted to the edge server for further processing. In this way, device-edge co-inference achieves a good balance among on-device computation and the communication overhead, and thus reduces the inference latency.



Existing works on device-edge co-inference mainly investigated model splitting \cite{JALAD}, while intermediate feature compression received less attention. 
Although DNNs can gradually abstract the intermediate feature layer by layer, the feature dimension may not decrease. There is an in-layer data amplification phenomenon \cite{ResNet}, i.e., the output data size of early layers may be larger than the original input data\cite{JALAD}. Thus, without effective feature compression, the model splitting point needs to be deep enough until the size of the intermediate feature is small enough, which leads to more layers deployed on the device and more on-device computation.
So the effectiveness of feature compression affects both the communication overhead and the amount of on-device computation.


In this paper, we propose an end-to-end trainable architecture, named BottleNet++, for efficient feature compression in \textit{device-edge co-inference} systems.
BottleNet++ consists of an encoder, a non-trainable channel layer, and a decoder. Both the encoder and decoder adopt lightweight convolutional neural networks (CNNs), which effectively act as joint source-channel coding \cite{jssc}.
Furthermore, by training the encoder and decoder in an end-to-end manner and explicitly considering the communication channel effect, BottleNet++ effectively exploits the fault-tolerant property of the DNN for a higher compression ratio. To further improve the model generalization ability in different channel conditions, the parameters of channel condition are considered by the encoder.

The main contributions of this paper are as follows:
\begin{itemize}
\item  We introduce an end-to-end trainable architecture, named BottleNet++, for efficient intermediate feature compression and transmission for device-edge co-inference.
\item  We design an encoder that can perform adaptive coding under different channel conditions, which is robust to the varying channel conditions and achieves graceful degradation of accuracy with a noisy channel.
\item Simulation results show that BottleNet++ achieves, with less than 2\% accuracy degradation, up to 64$\times$ bandwidth reduction on the additive white Gaussian noise (AWGN) channel and up to 256$\times$ bit compression ratio on the binary erasure channel (BEC) compared with the direct feature transmission without compression. With a higher compression ratio, BottleNet++ enables splitting a DNN at earlier layers, which leads to up to 3$\times$ reduction in on-device computation compared with other compression methods.
\end{itemize}


\section{Preliminary}
DNN has become a powerful method for many applications, but it typically requires a tremendous amount of computation. To deploy a DNN on source-constrained devices for edge AI applications, we consider hybrid deployment \cite{collaborative_intelligence}, called device-edge co-inference in this paper, which utilizes both the mobile device and the edge server for the execution to achieve a better tradeoff between communication and on-device computation. 
Our study focuses on increasing the compression ratio of the intermediate feature, which can reduce the communication overhead and on-device computation. It is achieved by exploiting the compression potential from model compression, joint source-channel coding, as well as the fault-tolerant property of neural networks.

\subsection{Network Splitting and Feature Compression}

Network splitting and feature compression are two critical problems for device-edge co-inference. 
The selection of the splitting point needs to reduce the latency due to on-device communication and feature transmission simultaneously.
Feature compression has attracted lots of recent attention. 
Many works have proposed different splitting and compression methods. The feature coding method proposed in \cite{Edge-host} applied JPEG coding and Huffman coding to compress the intermediate data. 
The method in \cite{pruning} combined network splitting and model pruning, which first prunes the weights of the network and then selects the splitting point based on the pruned model. However, the pruning step would consume plenty of time.
More recently, BottleNet\cite{bottlenet} proposed to encode the intermediate feature by a neural network and use a compression-aware training approach to reduce the accuracy loss. The name of our architecture, i.e., BottleNet++, is inspired by this work.
%

All of the works mentioned above only considered source coding, while assuming reliable communication over the wireless channel, i.e., they adopted the separate principle of source and channel coding.
Moreover, their design objective is to recover the intermediate feature at the edge server, either perfectly (with lossless source coding) or with tolerable distortion (with lossy compression).
But, for the device-edge co-inference systems, reliable communication is not necessary.
In other words, over-compression and inaccurate communication can be tolerated as long as they do not seriously affect the inference performance.
Motivated by the above discussion, we propose an end-to-end design approach based on joint source-channel coding, while exploiting the fault-tolerant property of DNNs. In the next two subsections, the two main ingredients, i.e., joint source-channel coding and the fault-tolerant property of DNNs, are introduced. The proposed framework will be presented in Section III.

\begin{figure*}[t]
\centerline{\includegraphics[width=19cm]{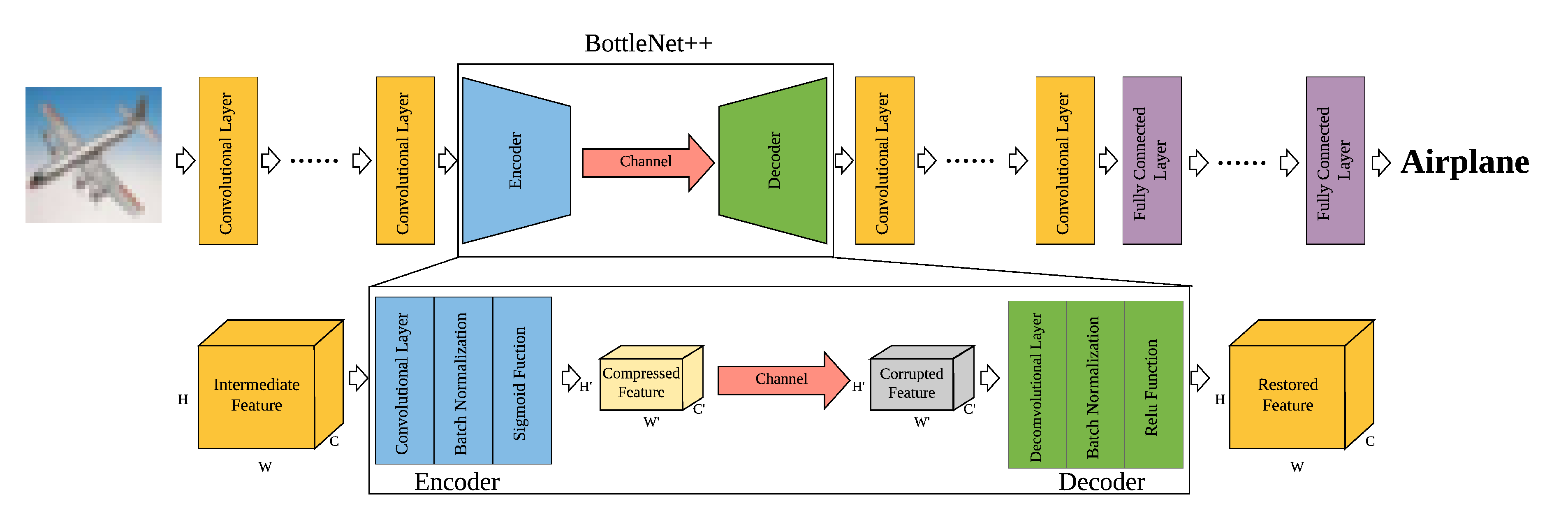}}
\caption{The architecture of BottleNet++.}
\label{architecture}
\end{figure*}

\subsection{Joint Source-Channel Coding} 
According to Shannon’s source-channel separation theorem\cite{shannon1948mathematical}, it is optimal to separate the design of source coding and channel coding. However, the optimality of the separate design holds only in the asymptotic limit of infinitely long source and channel blocks. In practice, joint source-channel coding can achieve better performance \cite{S4}.
Recently, many works have tried learning-based methods for joint source-channel coding. 
The work \cite{jssc} considered using the DNN to encode and decode the image over the AWGN channel.
Another work\cite{SS} implemented the joint source-channel coding on the text transmission over a binary erasure channel, which uses the sequence-to-sequence learning framework to encode and decode the text.

However, the above studies focused on restoring the transmitted message at the receiver, which is a challenging communication problem but is not fully aligned with the overall design objective in our considered problem.
As DNNs enjoy a fault-tolerant property, the high reliability in intermediate feature transmission is not needed, which gives further room for compression. To the best of our knowledge, this is the first work that exploits this unique opportunity for feature compression.

\subsection{Fault-Tolerance Property of Neural Networks}

Fault tolerance is frequently cited as a key property of neural networks\cite{1993} since these networks contain more neurons or processing elements than necessary to solve a given problem.
This property can be leveraged for reducing the communication overhead in the device-edge co-inference systems by relaxing the reliability requirement of transmission.
For this purpose, we propose to add the channel effect directly to the DNN during the training process, to exploit its fault-tolerant capability.
As to be shown in the experiments, thanks to the fault-tolerance property, even if the channel noise corrupts the transmitted data, the DNN performance degrades gracefully.

\section{Proposed architecture}

The proposed end-to-end architecture, namely, BottleNet++, is shown in the box of Fig. 1. It consists of an encoder, a non-trainable channel layer, and a decoder. BottleNet++ is deployed at the splitting point of DNN to compress and transmit the cubelike intermediate feature tensors.
The encoder and decoder are a pair of complementary lightweight CNNs. The extra computation introduced by the encoder and decoder is negligible compared to the whole network.
The wireless channel is modeled as a non-trainable layer in DNN and represented by a transfer function, similar to DNN based joint source-channel coding \cite{jssc}.
In this paper, we only consider two simple channel models, i.e., the AWGN and binary erasure channels. For different channel models, the transfer function can be replaced accrodingly, and our BottleNet++ can still work due to the network fault tolerance.

\subsection{Encoder}

The encoder plays the role of feature compression and joint source-channel coding, which consists of a convolutional layer, a batch normalization layer, and an activation layer. The encoder applies lossy compression to reduce the dimension of the intermediate feature of DNN. The convolutional layer uses different numbers of filters to control the output channel., and the stride of the convolutional operation and the kernel size determine the spatial size of the output feature.
Concretely, the intermediate feature can be represented by a tensor as (channel, width, height). The four cubes presented in Fig. \ref{architecture} are feature tensors with size $\left ( C, W, H \right )$ or $\left ( C{}', W{}', H{}' \right )$. To compress $C$ channels to $C{}'$ channels, the convolutional layer adopts $C{}'$ filters. To realize width-wise reduction from  $W$ to $W{}'$ and height-wise reduction from $H$ to $H{}'$, the stride of convolutional operation is set to $\left ( \left \lfloor W/W{}' \right \rfloor,\left \lfloor H/H{}' \right \rfloor \right )$.
The convolutional layer is followed by a batch normalization layer and a Sigmoid activation function to add non-linearity features in the encoder. We use a Sigmoid function as the activation function because the output values are constrained to $[0,1]$, which can be scaled to satisfy the transmitter output power constraint and further benefit to quantize data in the digital communication system.
Furthermore, because the wireless channel is time-varying, in order to improve the encoder generalization ability, the encoder is designed to be adaptive to different channel conditions.
Assuming the channel state information is available, we will add it as a parameter to the encoder.

\subsection{Channel Model}

After encoded, the compressed feature is sent through the `unreliable', which would be corrupted by the channel noise. This process is shown at the bottom box of Fig. \ref{architecture}.
We consider two different types of channel models. First we consider the AWGN model and the transfer function is written as $f(x) = x + n$, with $n\sim\mathcal{N}(0,\sigma^2)$.
The parameter $\sigma^2$ captures the noise variance as the channel condition. The analog communication scheme is adopted over the AWGN channel, i.e., symbols of the compressed feature will be directly modulated without digitizing or channel encoding.
Besides, we also consider the binary erasure channel (BEC), which is also adopted in \cite{SS} for a DNN-based communication system.
The BEC uses the bit erasure rate (BER) $p$ to model the deep fades or burst errors in the channel model, which has binary input and ternary output. 
The value of the erasure bit is set to the average of the bit taking 0 and 1 in our experiment. For instance, converting a binary number to a decimal number, the value of 110 is 6. However, if the channel erases the leftmost bit, the receiver would assign the average of 010 and 110, which is 4, to it.
For the BEC model, the continuous data must be quantized to $n$-bit string before transmission. The output of the Sigmoid function is constrained  in $[0,1]$, so the quantizer is $\widetilde{X}=round\left ( {X\cdot \left ( 2^{n}-1 \right )}\right )/\left (  {2^{n}-1}\right ) $, where the float-point $X$ is quantized to $\widetilde{X}$, presented by $n$-bit sequence.

The AWGN model can be used directly in the end-to-end training because the transfer function is differentiable. However, the transfer function of the BEC and the quantization are non-differentiable, hindering the back-propagation process. 
To solve this problem, we simply skip the channel model in the back-propagation.
In the experiment, we find that ignoring the quantization and channel corruption is feasible, and the model still converges when the value of $p$ is not too large.


\subsection{Decoder}

The decoder deployed at the receiver is a joint source-channel decoder to map the corrupted feature to the restored feature. It consists of a deconvolutional layer, a batch normalization layer, and a ReLU activation function.
The decoder aims to restore the bitstream to the feature tensor with the same dimension before compression.
The number of filters in the deconvolutional layer determines the output channel number. To recover $C{}'$ channels from $C$ channels, the deconvolutional layer use $C{}$ filters.
Width-wise and height-wise restoration can adopt $\left ( \left \lfloor W/W{}' \right \rfloor,\left \lfloor H/H{}' \right \rfloor \right )$ stride in convolutional operation, where the tuple value is the same as the encoder.
In our implementation, we set up the convolutional kernel size to 2$\times$2 and stride to $\left ( 2, 2 \right )$ in the encoder, which realize 2$\times$ width-compression and 2$\times$ height-compression, and the decoder also uses 2$\times$2 size kernel and $\left ( 2, 2 \right )$ stride.
For both the encoder and decoder, we use the convolutional/deconvolutional network to compress the intermediate feature but not a fully connected layer. This is because, although the fully connected layer has stronger compression capability, its memory and computation cost is unacceptable. 
Compared to traditional coding algorithms, e.g., the LDPC code \cite{LPDC} and the Huffman code, our method enjoys a higher computational efficiency, and the encoding and decoding models are much simpler.

\subsection{Training Strategy}

As illustrated in Section \uppercase\expandafter{\romannumeral3}.B,  the proposed BottleNet++ can be trained in an end-to-end manner with channel noise. 
However, directly training the whole architecture would suffer from the problem of slow convergence. Thus, we propose a three-step approach to train our end-to-end architecture.
The first step is to train the DNN, e.g., VGG16 \cite{VGG} or ResNet50 \cite{ResNet}, to reach the desired accuracy of the task. A DNN consists of many layers, like the top of Fig. \ref{architecture} excluding BottleNet++.
The second step is to select the splitting point to deploy the BottleNet++, and then train and update the weights of the encoder and decoder while fixing other parameters in the DNN.  In this step, training the encoder and decoder in different channel conditions, i.e., with different values of $\sigma$ in the AWGN channel and $p$ in the BEC, will benefit its generalization ability.
In the last step, we fine-tune the whole network to increase the accuracy further. In this process, all the parameters in both the DNN and BottleNet++ are updatable, and we train the whole BottleNet++ with a low learning rate.

\begin{figure*}[t]
\subfloat[]{
\begin{minipage}[]{0.242\linewidth}
\centering
\includegraphics[width=.97\textwidth]{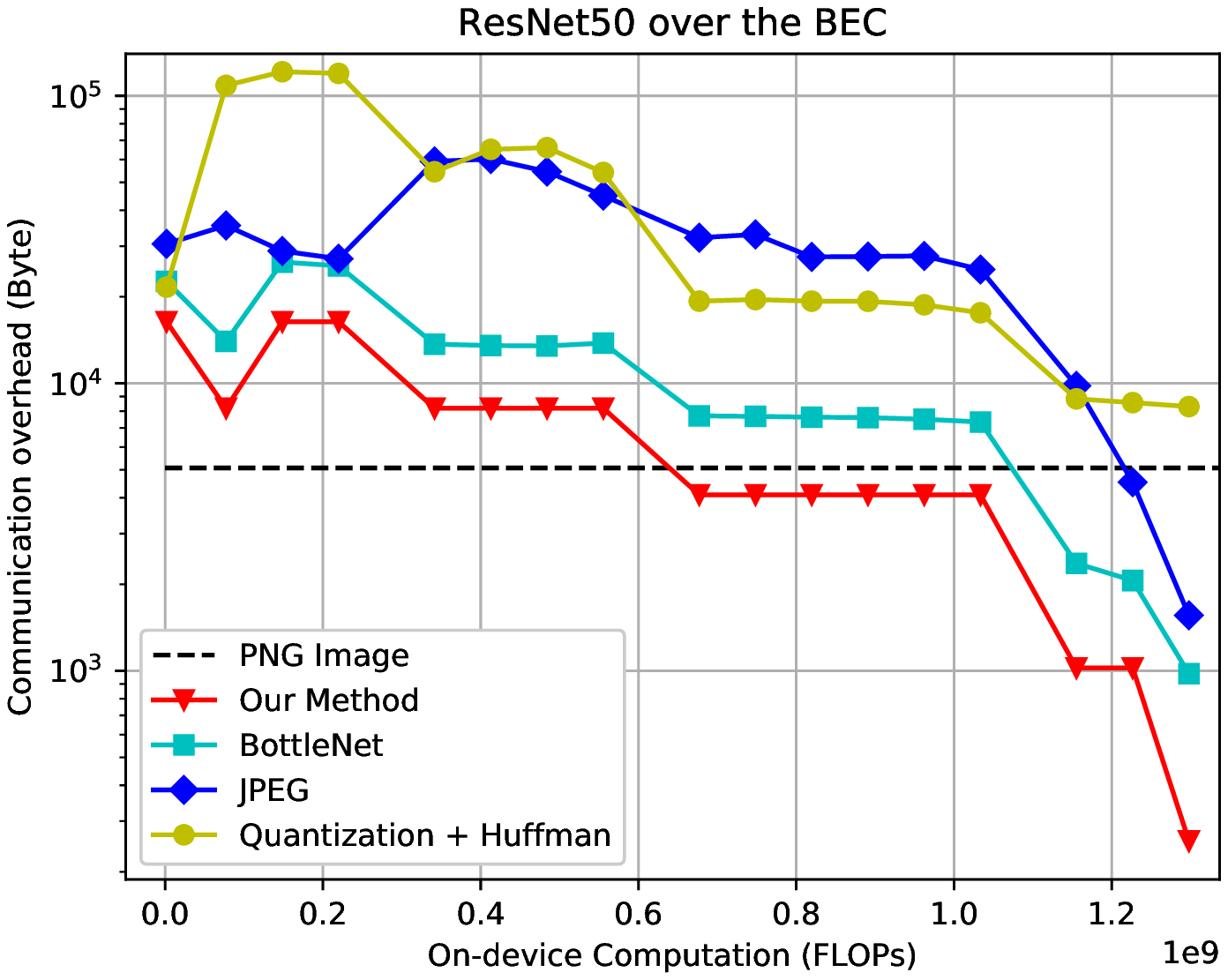}
\label{bec_resnet_compare}
\end{minipage}%
}%
\subfloat[]{
\begin{minipage}[]{0.242\linewidth}
\centering
\includegraphics[width=.97\textwidth]{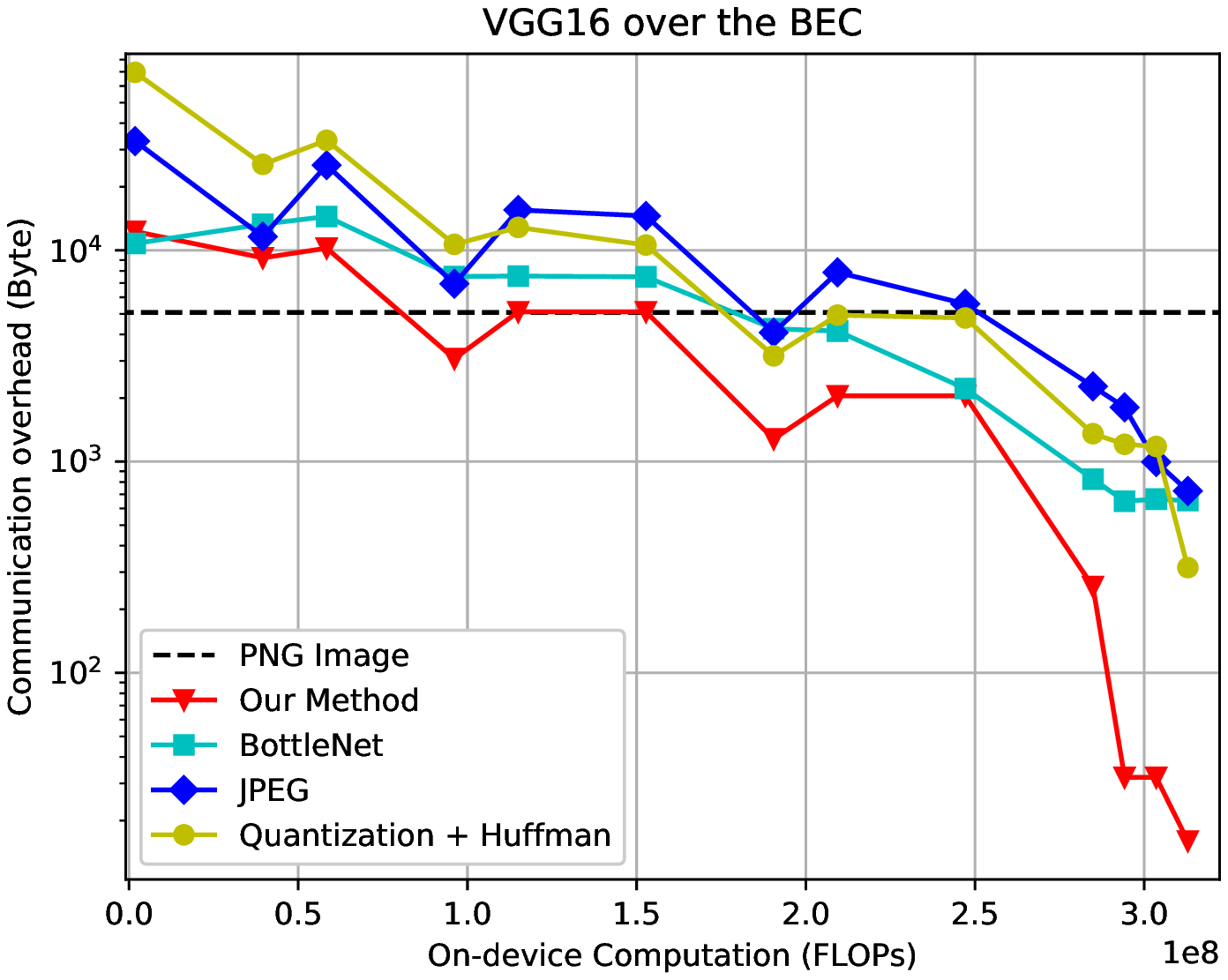}
\label{bec_vgg_compare}
\end{minipage}%
}%
\subfloat[]{
\begin{minipage}[]{0.242\linewidth}
\centering
\includegraphics[width=.97\textwidth]{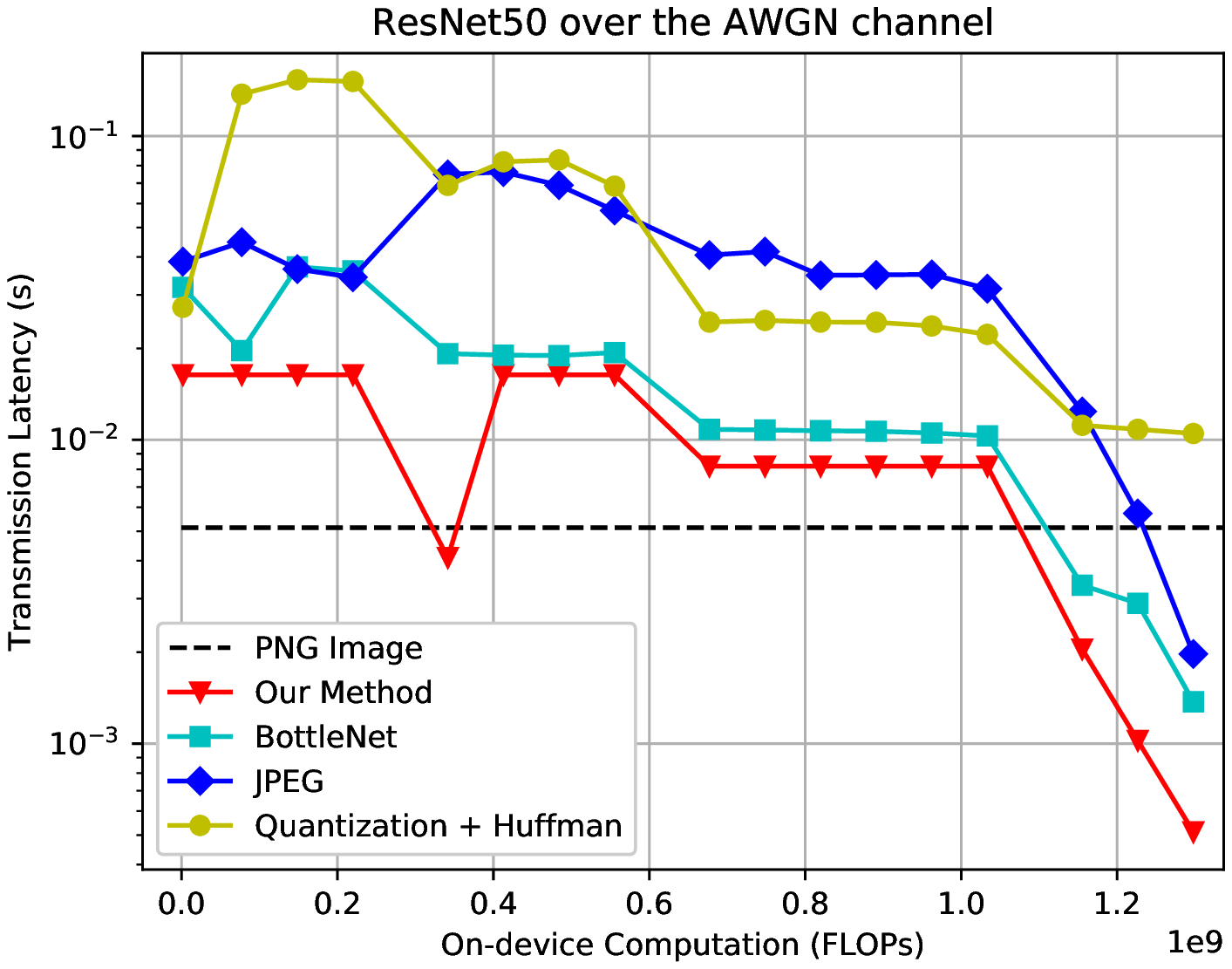}
\label{AWGN_resnet_compare_self}
\end{minipage}%
}%
\subfloat[]{
\begin{minipage}[]{0.242\linewidth}
\centering
\includegraphics[width=.97\textwidth]{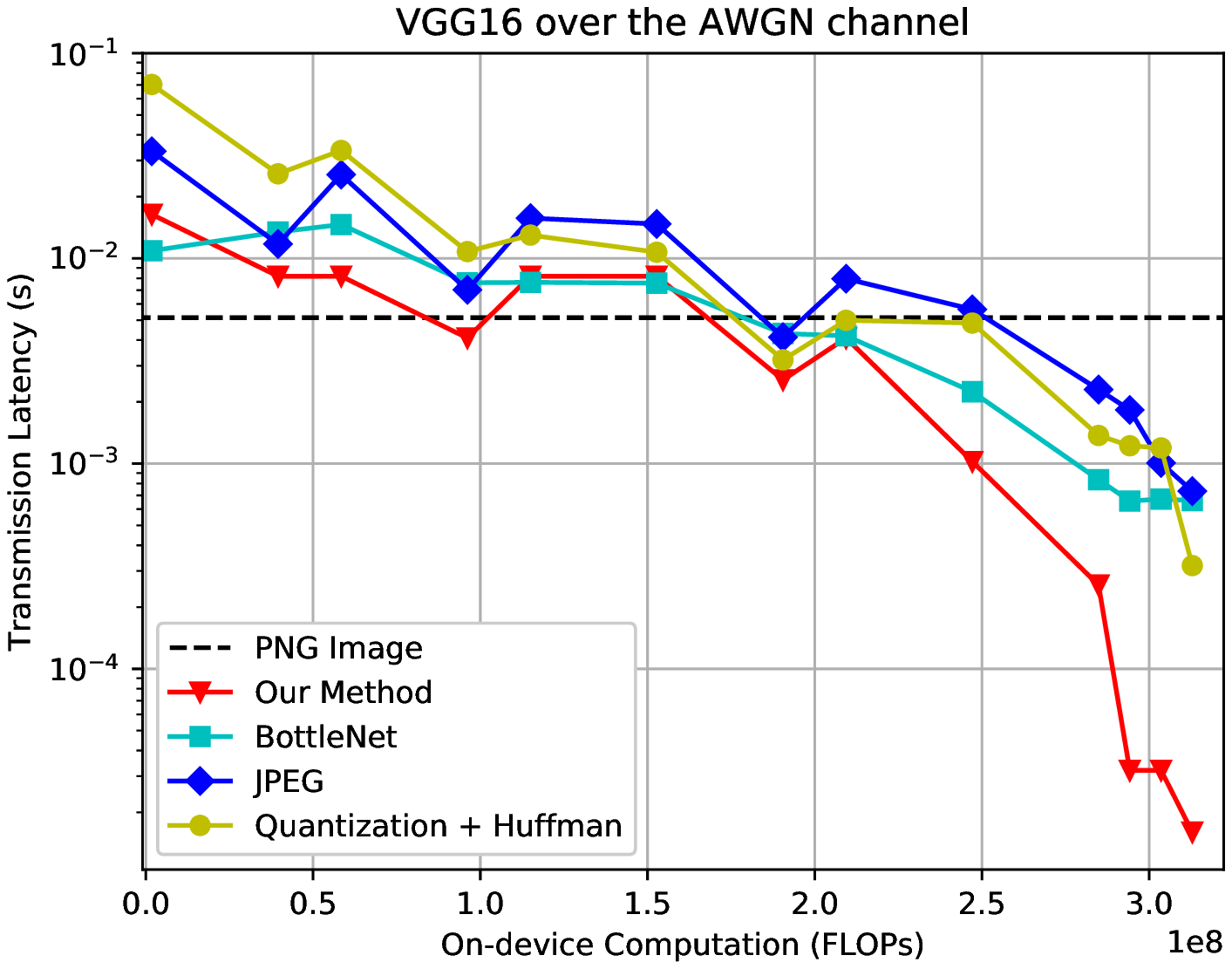}
\label{AWGN_vgg_compare_self}
\end{minipage}%
}%
\centering
\caption{On-device computation vs. communication overhead in BEC with (a) ResNet50 and (b) VGG16 and On-device computation vs. Transmission Latency in AWGN channel with (c) ResNet50 and (d) VGG16.}
\end{figure*}

\begin{table*}[htp]
\centering
\caption{minimum on-device computation in different scenes with the comminication overhead less than transmitting PNG image.}
\begin{tabular}{|c|c|c|c|c|}
\hline
On-device Computation (FLOPs) & BottleNet++ (ours)             & BottleNet      & JPEG           & Quan.+Huffman  \\ \hline
ResNet50 (BEC)      & $\mathbf{6.8\times 10^{8}}$ & $1.1\times 10^{9}$ & $1.2\times 10^{9}$ & $1.3\times 10^{9}$          \\ \hline
ResNet50 (AWGN)     & $\mathbf{3.4\times 10^{8}}$ & $1.1\times 10^{9}$ & $1.3\times 10^{9}$ & $1.3\times 10^{9}$              \\ \hline
VGG16 (BEC)         & $\mathbf{9.6\times 10^{7}}$ & $1.9\times 10^{8}$ & $1.9\times 10^{8}$ & $1.9\times 10^{8}$ \\ \hline
VGG16 (AWGN)        & $\mathbf{9.6\times 10^{7}}$ & $1.9\times 10^{8}$ & $1.9\times 10^{8}$ & $1.9\times 10^{8}$ \\ \hline
\end{tabular}
\label{tabel1}
\end{table*}

Compared with the architecture similar to our proposal, i.e., BottleNet \cite{bottlenet}, our method explicitly models the wireless channel as a non-trainable layer in DNN, while BottleNet assumes reliable communication over the wireless channel and ignores the bandwidth expansion caused by channel coding. As BottleNet++ considers channel conditions and exploits the fault-tolerance property of DNNs, the encoded bitstream does not need to be protected by a powerful channel encoder, and thus it enjoys a higher compression capability compared with BottleNet.
For the training process, BottleNet adopts a compression-aware training approach, which trains the DNN and the compression module simultaneously. In contrast, our method adopts a three-step approach to train the whole architecture, which improves the convergence rate. Moreover, when we need to adjust the compression ratio to different channel conditions, we only need to retrain the compression module instead of the whole model.


\section{Evaluation}

\subsection{Experimental Setup}

We consider a classification task with the CIFAR-100 dataset \cite{cifar100} that consists of 60,000 32$\times$32 color images in 100 classes, with 600 images per class.
Since the edge devices usually collect and deal with low-resolution images in IoT devices, CIFAR-100 is well suited for on-device applications.
Several DNN architectures have been proposed to achieve outstanding classification performance, e.g., VGGNet\cite{VGG}, ResNet\cite{ResNet}, DenseNet\cite{DenseNet}, etc. In this experiment, we use classical ResNet50 and VGG16. Although they can not achieve the start-of-the-art performance on CIFAR-100, our focus is on testing the compression capability of our method.
%
In our experiment, VGG16 reaches 74.04\% accuracy, and ResNet50 achieves 77.81\% accuracy. The configuration of ResNet50 approximately follows \cite{ResNet}, but, in the first convolutional layer, we change the kernel size to 3x3 and modify the stride of the convolutional operation to 1.
%
%
For VGG16, its structure directly follows \cite{VGG}.
The loss function in the training process is the cross-entropy function.

As discussed in Section III.B, we consider the AWGN channel and BEC. The peak signal-to-noise ratio (PSNR) is used to indicate the AWGN channel condition, and the bit erasure rate (BER) is used to indicate the quality of a BEC. Because the maximum output of the encoder is 1, we define PSNR as:
\begin{equation}
\mathrm{PSNR} = 10\log_{10}\frac{1}{\sigma ^{2}}\;\; (\mathrm{dB})
\end{equation}

To evaluate our proposed method, we split the DNN and compress the intermediate feature at different splitting points of VGG16 and ResNet50.
Note that not all the layers in a DNN can be used as a splitting point. For sequential DNNs like VGG and AlexNet\cite{alexnet}, input signals flow layer by layer, and we can easily split the network at the end of each layer. However, the latest deep models like ResNet introduce branchy network structures rather than sequential models. So, the splitting points are different. In our evaluation, each res-unit in ResNet\cite{ResNet} is regarded as a possible splitting point, while for VGG16, each convolutional layer can be regarded as a splitting point.
The code is available at github.com/shaojiawei07/BottleNetPlusPlus.
%




\subsection{Compression Capability Comparison}
The latency of the device-edge co-inference system is mainly composed of on-device computation latency and transmission latency/communication overhead. So, we should split the network as early as possible (close to the input layer), with less on-device computation, and compress the intermediate feature as much as possible without too much loss of accuracy. In the experiment, the accuracy degradation threshold is set to 2\%, and the on-device computation cost is approximated by the number of floating-point multiplication operations (FLOPs) in the convolutional layers before the splitting point.

In the following, we compare the compression capability of BottleNet++ with other methods, in both the BEC and AWGN channels, at different splitting points.  
Three baseline methods are considered: the method in \cite{Edge-host} that adopts the JPEG algorithm for lossy compression, denoted as ``JPEG''. The approach in \cite{JALAD} that quantizes floating-point data to $n$ bit-depth and then encodes the result with Huffman coding denoted as ``Quantization + Huffman'', and BottleNet\cite{bottlenet} that encodes the intermediate feature by a neural network and uses JPEG compression-aware training to reduce the accuracy loss. 
Furthermore, we consider the communication overhead of transmitting the raw PNG image from device to edge, denoted as ``PNG image".
All of the baseline methods assumed reliable communication with necessary channel coding.
In contrast, because our method integrates the channel model in the neural network, it can avoid extra cost for channel coding.
We first conduct the experiment over the BEC with the bit erasure rate $p=0.01$.
To ensure fairness, we apply the 1$/$2 rate LDPC code \cite{LPDC} to other methods as channel coding. Different code rates may affect the baseline curves in Fig. 2, but they do not change the result in Table \ref{tabel1}. Extensive experiment results show that even when the channel conditions and the code rates change, our BottleNet++ still performs better than other baseline methods and has similar results as Table \ref{tabel1}.
%
The compression capability is represented by the communication overhead (size of the transmitted bitstream), which is equivalent to transmission latency.
The result is shown in Fig. \ref{bec_resnet_compare} and Fig. \ref{bec_vgg_compare} and we note that almost at any splitting point, BottleNet++ achieves the lowest communication overhead. It realizes up to 256$\times$ bit compression ratio in the last convolutional layer of ResNet, where BottleNet++ compresses 8192 32bit-floating numbers (32 KB) to 128 8bit-integers (128 Bytes).
%
%

Besides, with the requirement that the communication overhead of the intermediate feature should be less than that of raw PNG image, our BottleNet++ can split the network earlier than other methods.
This result is summarized in Table \ref{tabel1}, which shows the on-device computation of different methods at the earliest splitting point. 
In any case, our BottleNet++ achieves the minimum on-device computation.

We next investigate the performance of compression capability over the AWGN channel.
In this case, we directly use the transmission latency as the indicator.
Compared with baseline methods, BottleNet++ adopts the analog communication for transmitting the compressed feature, which bypasses quantization, source coding, and channel coding, and directly maps the data to continuous samples before transmission. We use quadrature modulation, and each of the I and Q channels carries an encoded but unquantized symbol.
As the baseline methods assume perfect channel coding, we adopt the Shannon capacity formula to calculate their data rate, i.e., $C=W\log_2(1+\mathrm{SNR})$.
Note that this gives a performance upper bound for the baseline methods, and thus in practice, the performance gain of BottleNet++ will be more prominent.
%
The experiment fixes $W=1$ $\mathrm{MHz}$, $\mathrm{SNR}=14.5$ $\mathrm{dB}$, and adopts transmission latency to evaluate the communication overhead. Specially, we set $\mathrm{PSNR}=20$ $\mathrm{dB}$ in BottleNet++ to satisfy its SNR equal to 14.5 $\mathrm{dB}$, which is calculated from the transmitting power of each encoded symbol.

As shown in Fig. \ref{AWGN_resnet_compare_self} and Fig. \ref{AWGN_vgg_compare_self}, even assuming perfect coding for digital communications, the baseline methods introduce higher transmission latency than BottleNet++.
In particular, BottleNet++ achieves up to 64$\times$ bandwidth reduction in the second last convolutional layer of ResNet50, where it compresses the 2048-symbol feature to 32 symbols.
Furthermore, similar to the BEC case, Table \ref{tabel1} shows that BottleNet++ reduces on-device computation by $\sim$2$\times$ and $\sim$3$\times$ for VGG16 and ResNet50 in the AWGN channel compared with other methods.

\begin{figure}[t]
\centering
\subfloat[]{
\begin{minipage}[t]{.235\textwidth}
\centering
\includegraphics[width=0.99\textwidth]{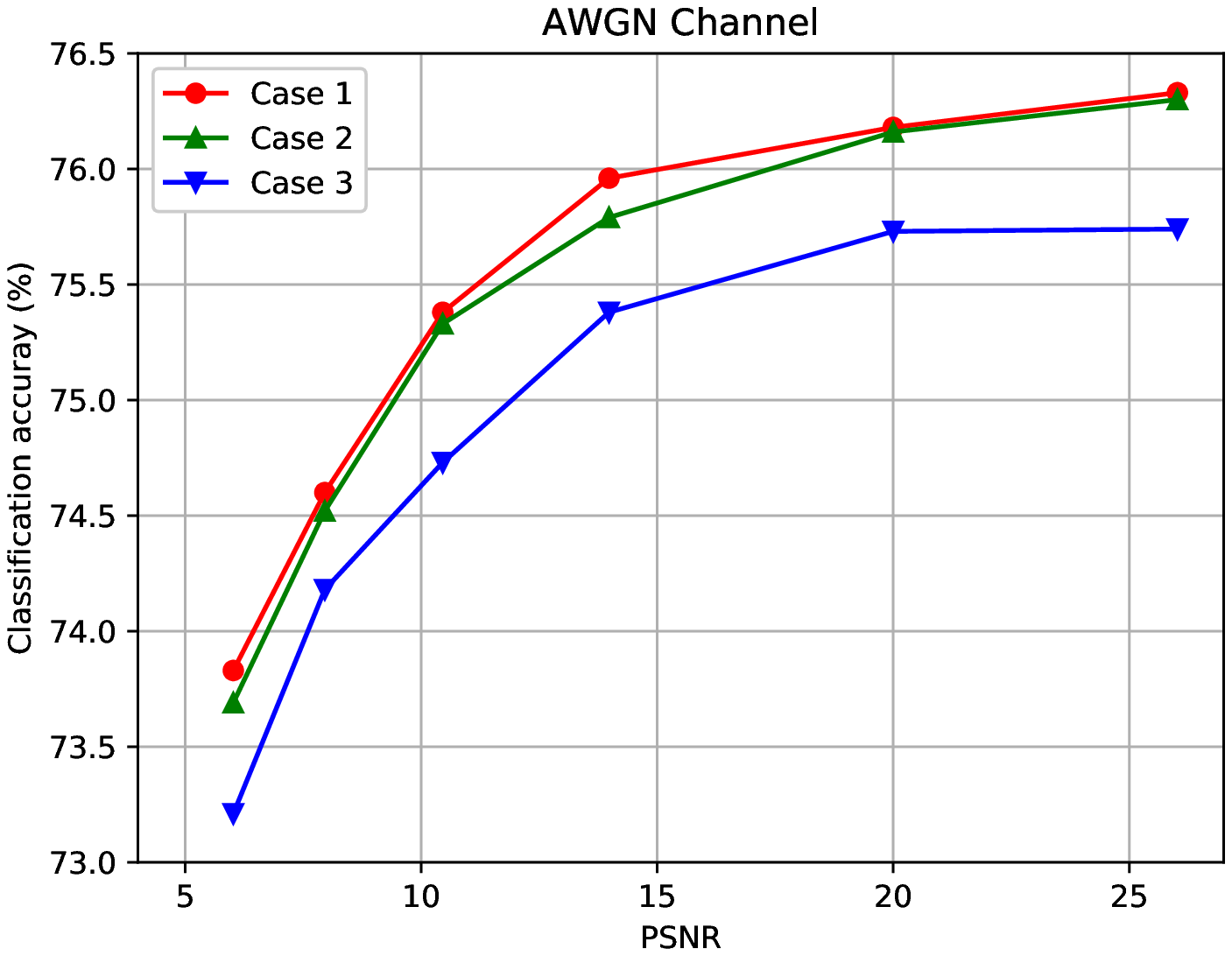}
\label{awgn_degradation}
\end{minipage}%
}%
\subfloat[]{
\begin{minipage}[t]{.235\textwidth}
\centering
\includegraphics[width=0.99\textwidth]{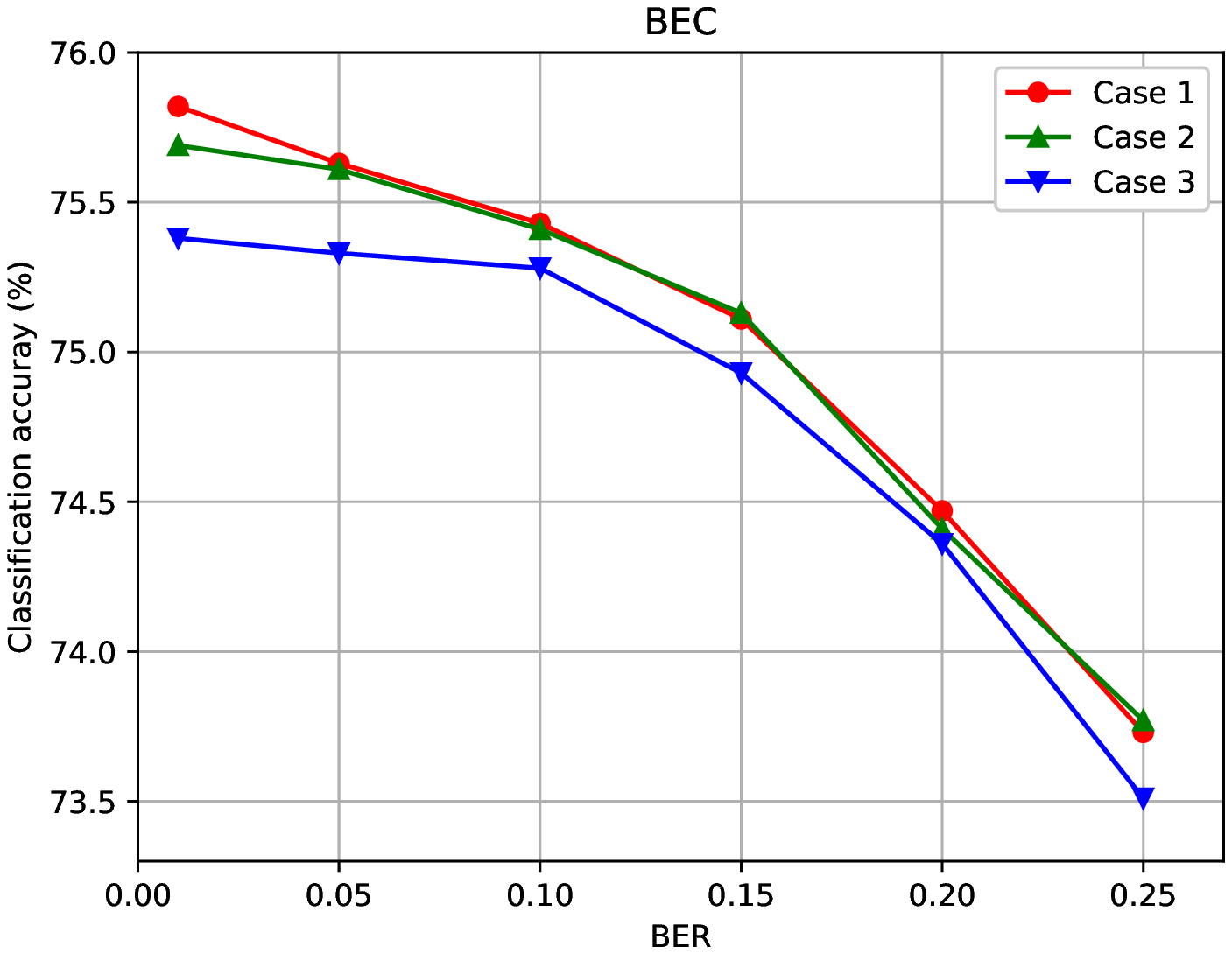}
\label{bsc_degradation}
\end{minipage}%
}%
\caption{Accuracy degradation of BottleNet++ over the (a) AWGN channel and (b) BEC.}
\end{figure}




\subsection{Generalization Ability and Robustness Analysis}

In real communication systems, the channel condition is time-varying, and in this part, we test the generalization ability, i.e., accuracy degradation, of BottleNet++ when the channel condition changes.
We conduct the experiment based on the ResNet50 model deployed behind the last convolutional layer, with 64$\times$ compression ratio for the AWGN channel and with 256$\times$ bit compression ratio for BEC, respectively. 
We evaluate the performance of BottleNet++ in three different cases: 
\begin{itemize}
\item Case 1: The encoder knows the channel state information, i.e., the bit erasure rate or PSNR, in both the training and testing processes.
\item Case 2: The encoder only knows the channel state information in the training process. In the testing process, the encoder assumes the channel condition to be 15 $\mathrm{dB}$ in the AWGN channel or 0.125 in the BEC.
\item Case 3: The encoder does not know the channel state information in either the training or the testing process.
\end{itemize}

Fig. \ref{awgn_degradation} and Fig. \ref{bsc_degradation} presents the accuracy loss in different channel conditions of the three cases. Case 1 achieves the highest accuracy under any channel condition. Case 2 is very close to case 1, which shows the robustness of BottleNet++ to the variation of channel conditions. Case 3 has a noticeable accuracy drop compared to Case 1, which means that considering channel conditions during encoding can improve the generalization ability.
Remarkably, the performance of BottleNet++ is robust to channel variations in all three cases. Specifically, the accuracy drops less than 1\% when PSNR changes from 25 $\mathrm{dB}$ to 10 $\mathrm{dB}$ in the AWGN channel, or when the bit erasure rate changes from 0.01 to 0.15 in BEC.



\section{Conclusions}

In this paper, we proposed an end-to-end deep learning architecture, named BottleNet++, for device-edge co-inference with resource-constrained mobile devices.
By exploiting the strong sparsity and the fault-tolerant property of the intermediate feature in the DNNs, BottleNet++ achieves a much higher compression ratio than existing methods, which leads to a significant reduction in the communication overhead, and makes it feasible to split a DNN in the earlier layer to reduce the on-device computation.
For the communication theoretic aspect, our study casts new light on the two fundamental problems in the setting of device-edge co-inference: \emph{What to transmit}? \emph{How to transmit}? The results indicate that transmitting highly compressed features with analog communication becomes attractive for edge-assisted inference. Furthermore, DNNs stand out as powerful design tools for such new communication problems.


\bibliographystyle{./IEEEtran}
\bibliography{IEEEabrv,ref}

\end{document}